%% file: main.tex
\documentclass{article}

\PassOptionsToPackage{numbers, compress}{natbib}

\usepackage[final]{neurips_robustseq_2022}

\usepackage[utf8]{inputenc} 
\usepackage[T1]{fontenc}    
\usepackage{hyperref}       
\usepackage{url}            
\usepackage{booktabs}       
\usepackage{amsfonts}       
\usepackage{nicefrac}       
\usepackage{microtype}      
\usepackage{xcolor}         
\usepackage{graphicx}

\usepackage{lscape}         

\input{tex/variable_definitions.tex}
\everypar{\looseness=-1}    %
\title{Probabilistic thermal stability prediction through sparsity promoting transformer representation}

\author{%
  Yevgen Zainchkovskyy\thanks{Corresponding author} \\
  DTU Compute \& Novo Nordisk A/S \\
  \texttt{yezs@novonordisk.com}
  \And
  Jesper Ferkinghoff-Borg \\
  Novo Nordisk A/S\\
  \texttt{jfgb@novonordisk.com}
  \And
  Anja Bennett \\
  Novo Nordisk A/S \\
  \texttt{zabx@novonordisk.com}
  \And
  Thomas Egebjerg \\
  Novo Nordisk A/S \\
  \texttt{tegb@novonordisk.com}
  \And
  Nikolai Lorenzen \\
  Novo Nordisk A/S \\
  \texttt{nlz@novonordisk.com}
  \And
  Per Jr.~Greisen \\
  Novo Nordisk A/S \\
  \texttt{pjug@novonordisk.com}
  \And
  S\o ren Hauberg \\
  DTU Compute \\
  \texttt{sohau@dtu.dk}
  \And
  Carsten Stahlhut \\
  Novo Nordisk A/S \\
  \texttt{ctqs@novonordisk.com}
}

\begin{document}

\maketitle

\begin{abstract}
Pre-trained protein language models have demonstrated significant applicability in different protein engineering task \cite{rao2019evaluating,rives2019biological}. A general usage of these pre-trained transformer models producing latent representation is to use a mean pool across residue positions to reduce the feature dimensions to further downstream tasks such as predicting bio-physical properties or other functional behaviours. In this paper we provide a two-fold contribution to machine learning (ML) driven drug design. Firstly, we demonstrate the power of sparsity by promoting penalization of pre-trained transformer models to secure more robust and accurate melting temperature (Tm) prediction of single-chain variable fragments with a mean absolute error of $0.23^{\circ}$C. Secondly, we demonstrate the power of framing our prediction problem in a probabilistic framework. Specifically, we advocate for the need of adopting probabilistic frameworks especially in the context of ML driven drug design.
\end{abstract}

\input{tex/introduction.tex}
\input{tex/background.tex}
\input{tex/idea.tex}
\input{tex/results.tex}

\input{tex/conclusions.tex}

\begin{ack}
This project was supported in part by the Innovation Fund Denmark via the Industrial PhD Programme (grant no.\@
0153-00007B), Novo Nordisk R\&D STAR Fellowship Programme and the Novo Nordisk Foundation (grant no.\@
NNF20OC0062606) via the Center for Basic Machine Learning Research in Life Science (MLLS, \url{https://www.mlls.dk}).
\end{ack}

\bibliographystyle{unsrt}
\bibliography{main}
\newpage
\input{tex/appendix.tex}
\end{document}

%% file: tex/variable_definitions.tex
\newcommand\f{\mathbf{f}}
\newcommand\x{\mathbf{x}}
\newcommand\X{\mathbf{X}}
\newcommand\y{\mathbf{y}}

\newcommand\Z{\mathbf{Z}}
\newcommand\K{\mathbf{K}}
\newcommand\I{\mathbf{I}}

%% file: tex/introduction.tex
\section{Introduction}\label{sec:Intro}
Peptide and protein engineering is the process of optimizing peptide and proteins towards desired and valuable features for technological or medical applications \cite{bloom2006protein}. In protein engineering, we seek to optimize the function of a protein with respect to e.g. its expression level, solubility, or thermal stability. Their functional behavior is directly determined by their amino acid sequence. Thus, to develop new or optimize desired properties for e.g., biomedical applications require to invert the relationship of the function given the sequence \cite{cocco2018inverse}, also generally known in statistics and machine learning as the inverse problem. However, existing design methods have serious problems in distinguishing the functional levels of closely related proteins \cite{dou2017sampling,yang2019machine}. While both protein engineering and design is a NP-hard problem \cite{pierce2002protein}, a direct search in the protein space simply becomes an overwhelming and intractable approach in linear time. Directed evolution has successfully demonstrated its applicability of mapping peptide and protein sequencing to functional behavior. However, it is highly limited by the fact that even high-throughput techniques only can sample a minor fraction of sequences constructed from diversification methods \cite{tokuriki2009stability}. Among others, Bedbrook and co-workers have demonstrated the direct applicability of utilizing machine learning for optimizing a property that would not have been possible to engineer through directed evolution alone, \cite{yang2019machine, bedbrook2019machine, khare2011baker}. On the other hand, machine learning models are heavily dependent on learning from data - a crucial part in designing a machine learning driven drug design pipeline is therefore the accessibility of relevant functional data for the task at hand.

In this contribution, we demonstrate and discuss classical challenges in both designing compounds with dedicated properties from a minimal set of observations and data sets with quite scarce diversity. While we at one hand wish to provide as diverse molecules to maximize the coverage of the chemical search space and on the other hand seek to ensure optimized properties within minimum number of design rounds (experiments) - we are facing a typical active learning problem balancing explore vs exploit steps through the usage of model uncertainty estimates. In section~\ref{sec:Background}, we provide a unified probabilistic framework for integrating compact latent pre-trained transformer features with Gaussian Process (GP) regression models, \cite{rasmussen2003gaussian}. Through careful variant design train, development (dev), and test splits we demonstrate the applicability of uncertainty estimates to assess the models own notion of what it does not know. We examine the effect of training data with 1-5 mutations away from a wild-type sequence and the models ability to reason of its own predictive power to generalize to multiple mutations. While we in this paper limit ourselves to the quantification of predictive performance of the models, the GPs can be utilized as the surrogate model in a Bayesian Optimization framework for optimizing and searching the sequence space.

%% file: tex/background.tex
\section{Background}
\label{sec:Background}
Here, we motivate our problem and provide a brief overview of the core architecture of our probabilistic models utilizing a transformer architecture as input to our downstream regression models. Significant improvement and applicability of protein language models have been demonstrated over the last years, where among others the UniRep \cite{alley2019unified}, Evolutionary Scale Modeling (ESM) \cite{rives2019biological}, ProtBert \cite{elnaggar2021prottrans} models can be mentioned. In \cite{alley2019unified}, they utilize pretrained language model representation UniRep to generalize the representation to unseen regions of sequence space.
Furthermore, Vig \& Rao argues for attention in the transformer models corresponds to known biological properties like structure and binding sites that can enable contact prediction \cite{vig2020bertology,rao2020transformer}.
In a drug design setting, we are especially interested in enabling pretrained representations in our protein engineering tasks for designing improved drugs as we are highly limited by the number of experiments we can conduct relative to the enormous sequence space at hand, e.g. $10^{130}$ for proteins of 100 amino-acids length. Thus, searching the space intelligently is needed even when utilizing high-throughput experimental setups.

While designing or engineering proteins, we are faced with the problem of optimizing towards specific functions of the molecules and effectively only interested in a tiny subspace of sequence space. The main challenge is naturally how can we utilize the general protein representation for a direct fine tuning to the downstream functional optimization task at hand. In this contribution, we seek to build a model for predicting the thermal stability of the antibody format single chain variable fragment (scFv). Due to the small sizes and the stranded nature of scFvs, these are commonly used as building blocks to construct recombinant multi-specific antibody formats, \cite{bird1988single}. Unfortunately, the scFvs has been reported to be less termostable than larger antibody formats and thus more likely to lead to undesired aggregation and low Tms when utilized in a multi-specific format, \cite{fink1998protein}. To improve biophysical behavior of the scFV, our goal is to build a predictive model of the experimental measured Tm values determined by nano differential scanning fluorimetry (nanoDSF) \cite{strutz2016exploring}. Having an accurate model for prediction the melting temperature is needed to assess which variants to test experimentally for increased thermal stability. To quantify the accurate of our predictions, we follow a probabilistic approach where we not only obtain our mean predictions but just as importantly can provide uncertainty estimates on the predictions. Uncertainty estimates is critically needed for providing quantitative and directed search strategies balancing both exploitation and exploration.

\subsection{Transformer-based models}
Transformers are revolutionising NLP, have recently been repurposed to model biological sequences.In the core of a transformer architecture is the attention mechanism allowing to capture long-range dependencies between positions in a sequence. Originating as a solution to classic sequence-to-sequence (seq2seq) models, attention mechanism shows better performance and scaling characteristics than traditional RNNs or LSTMs. Common to those architectures is a context vector comprising of a hidden state of the network being carried through subsequent propagation, resulting in degraded performance with increased sequence lengths. On the other hand, transformers utilize self-attention, which allows processing of the whole sequence while still focusing on specific parts of it. 

In the context of biological sequence modelling, the hidden state of a Transformer model corresponds to individual amino acid residues and represents the given amino acid in its context as a point in a high dimensional space (embedding). Thus, similar sequences are assigned similar representations by the network and are mapped to nearby points in space.

In this contribution, we use the ESM1-b variant of a Transformer protein language model from Facebook AI Research \cite{rives2019biological} encoding each of our sequences to an embedding $\x \in \mathbb{R}^{250\times1280}$.

\subsection{Gaussian process regression}
A Gaussian Processes (GP) is a powerful probabilistic framework enabling nonparametric, nonlinear Bayesian models \cite{rasmussen2003gaussian}. A GP defines a prior distribution over the set of function $f\left( \x \right)$ mapping the relation between our $M$-dimensional feature representation of our protein sequences to the target property $y = f\left( \x \right) + \epsilon$. Here $\epsilon$ represents additive observation noise. Using a standard zero-mean GP prior we obtain
\begin{equation}\label{eq:GPzeroMeanPrior}
    p \left( f \left( \X \right) \right) = p \left( \f_X \right) = {{\mathcal{N}}\left( 0, \K \right)},
\end{equation}
where $\K$ is the covariance matrix between our training input features $\X = [\x_1, \cdots, \x_N]$ such that $K_{ij} = k\left( \x_i, \x_j \right)$ defines the covariance function between input $\x_i$ and $\x_j$.
We utilize one of the typically applied kernels for GP regression, Matern $\frac{5}{2}$ covariance function, with shared length-scale parameters for each input dimension $\sigma_l$, yielding
\begin{equation}\label{eq:Matern52Kernel}
k\left(\x_i, \x_j\right)=\sigma_f^2\left(1+\frac{\sqrt{3} r}{\sigma_l}\right) \exp \left(-\frac{\sqrt{3} r}{\sigma_l}\right) \quad \mathrm{where} \quad r=\sqrt{\left(\x_i-\x_j\right)^\top\left(\x_i-\x_j\right)}
\end{equation}
Assuming additive independent identically distributed Gaussian noise with variance $\sigma^2_\epsilon$ our predictive distribution for our new test proteins $\Z$ reads, \cite{rasmussen2003gaussian},
\begin{eqnarray}
    p\left( f_Z  | \X, \y, \Z \right) &=& \mathcal{N} \left( \mathbf{\mu}_z , \mathbf{\Sigma}_z \right), \quad \mathrm{where} \quad
    \mathbf{\mu}_z = k\left( \Z, \X \right)\left( \K + \sigma^2_\epsilon \I \right)^{-1} \y \nonumber \\
    \mathbf{\Sigma}_z &=& k\left( \Z, \Z \right) - k\left( \Z, \X \right)\left( \K + \sigma^2_\epsilon \I \right)^{-1} k\left( \X, \Z \right).
\end{eqnarray}

%% file: tex/idea.tex
\section{Driving sparsity through learned masks}

Having extracted an embedding for each residue in a sequence of length $P$, a typical approach used to represent a complete protein as a single vector $\hat \mathbf{x}$ is by averaging across the transformer's hidden representation $\mathbf{x}$ at each sequence position $p$ (mean pooling):
\begin{equation}\label{eq:MeanPoolVanilla}
\hat \mathbf{x} = \frac{1}{P} \sum\nolimits_{p=1}^{P}{\mathbf{x}_p}
\end{equation}
While significantly reducing the overall dimensionality of the embedding and allowing to represent proteins of different lengths, this approach inevitably results in loss of information. Intuitively, averaging assigns equal weight to all residues in the sequence, while in reality, only a handful of positions might influence the target of interest. 

In this work, we investigate 3 different positional-weighted approaches to the typical averaging: a positively constrained \textbf{Learned mask} $\mathbf{w_{l}}$, sparsity promoting \textbf{Sigmoid-transformed mask} $\mathbf{w_{s}}$ and a Half-Cauchy \textbf{Prior based mask} $\mathbf{w_{p}}$:

\noindent
\begin{minipage}{.34\linewidth}
\begin{equation}
\mathbf{\hat x_l} = \frac{ \sum_{p=1}^{P}{\exp({\mathbf{w_l}_p}) \mathbf{x}_p} } {\sum{\exp({\mathbf{w_l})}}}
\end{equation}
\end{minipage}%
\begin{minipage}{.66\linewidth}
\begin{equation}
\mathbf{\hat x_s} = \frac{ \sum_{p=1}^{P}{\mathrm{S}({\mathbf{w_s}_p}) \mathbf{x}_p} } {\sum{\mathrm{S}({\mathbf{w_s})}}} \quad \textrm{where} \quad \mathrm{S}(x) = \frac{1}{1 + \exp(-x)}
\end{equation}
\end{minipage}

\begin{equation}
\mathbf{\hat x_p} = \frac{ \sum_{p=1}^{P}{{\mathbf{w_p}_p} \mathbf{x}_p} } {\sum{{\mathbf{w_p}}}} \quad \textrm{where} \quad \mathbf{w_p} \sim \textrm{Half-Cauchy}(0, \sigma)
\end{equation}

From a practical perspective, for all three approaches, we learn the masks as a part of the standard gradient based GP Maximum Log-Likelihood maximization procedure.

%% file: tex/results.tex
\section{Results}

Using Mean Absolute Error (MAE) as our metric, and partitioning the data set, we ran 3 sets of experiments corresponding to 3 different splits. First, we tested on the subset of the training set where single-site mutations were used as the validation (1MUT), next we used a random sample of the training set as validation (Uniform Shuffle) and finally evaluated the proposed methods on the hold-out test set itself. The reason for this partitioning is the spread of the positions of mutations in the wild-type sequence and relative sizes of training and validation set. Naturally, having learned a specific mask on the smaller training set, performance will degrade if mask does not reflect the mutated positions in the bigger test-set. This effect is simulated for the "Uniform Shuffle" split where the size of validation set was 24 samples (vs. 10 samples for the 1-MUT split). 

\begin{table}[h]
\caption{Evaluation of the Baseline and the proposed methods. For each split and method we report mean $\pm$ std over 64 trials. Bold values denote statistical significance against Baseline ($p < 0.05).$}
\label{tab:results} 
\centering
\begin{tabular}{llll}
\toprule
 & 1-MUT Shuffle & Uniform Shuffle & Test Set (Hold out) \\
\midrule
Baseline               & \,\,1.216 $\pm$ 0.306 & 0.819 $\pm$ 0.157 &\,\,0.273 $\pm$ 0.006 \\
Learned mask           & $\mathbf{1.110 \pm 0.338}$ & 0.836 $\pm$ 0.231 & $\mathbf{0.227 \pm 0.010}$ \\
Learned mask (Sigmoid) & $\mathbf{1.150 \pm 0.347}$ & 0.813 $\pm$ 0.161 & $\mathbf{0.227 \pm 0.008}$ \\
Learned mask (Prior)   & $\mathbf{1.090 \pm 0.314}$ & 0.841 $\pm$ 0.158 & $\mathbf{0.226 \pm 0.010}$ \\
\bottomrule
\end{tabular}
\end{table}

Gross metrics are reported in Table~\ref{tab:results}. Here, we see that the \textbf{Learned Mask (prior)} proposed method outperforms the standard averaging approach referenced as \textit{Baseline} in the cases of balanced train/validation splits. Naturally, this method also depends on the actual value of the prior: $\sigma$ - the second moment of Half-Cauchy distribution. In our case, a prior of $\sigma=0.15$ was chosen as a result of a parameter sweep on the 1-MUT Shuffle resulting in lowest MAE. That prior is then reused for all \textbf{Learned Mask (prior)} runs. 

\begin{figure}[h]
\centering
\vspace{-0.5em}
\includegraphics[width=1.0\linewidth]{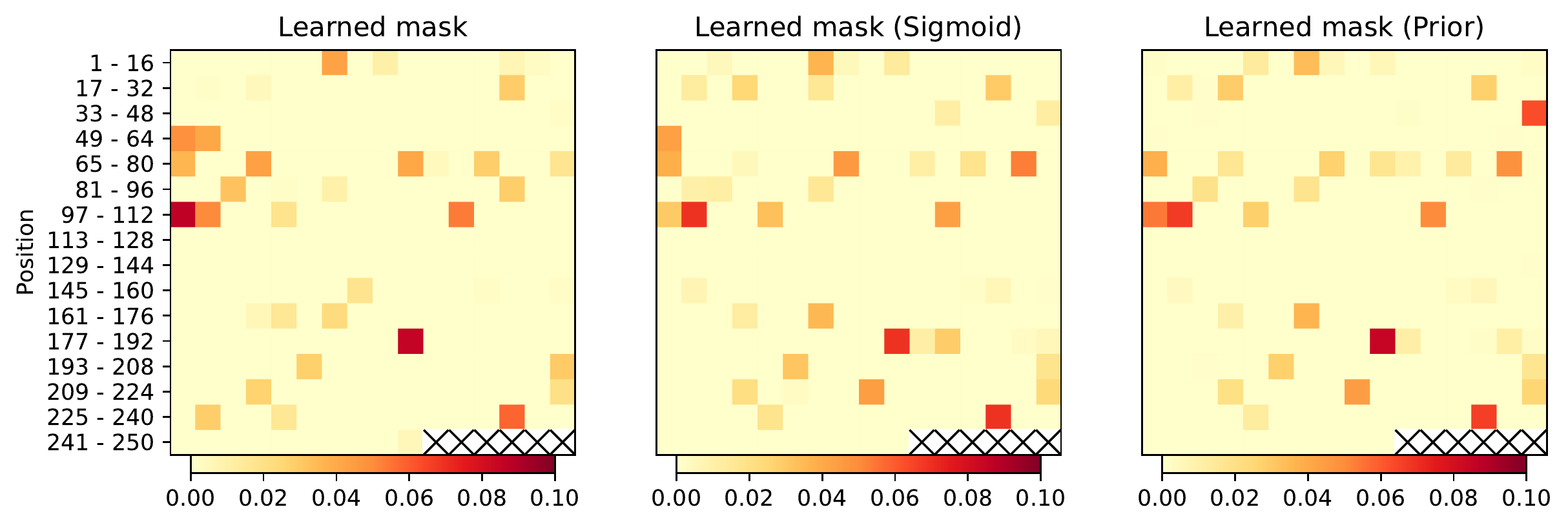}
\vspace{-1em}
\caption{Masks learned on the hold out set. The number of zero entries (values below $1e^{-5}$ threshold) are: \textbf{208 for Learned mask}, \textbf{203 for Learned mask (Sigmoid)} and \textbf{9 for Learned mask (Prior)}. Unused entries are marked with a cross $\times$.}
\label{fig:masks}
\end{figure}

Learned masks are shown in Figure~\ref{fig:masks}. In terms of the importance on melting temperature they emphasize roughly same positions in the sequence. Interestingly, the addition of Half-Cauchy prior, results in a more \emph{dense} mask, as seen on the number of zero-entries. This appear to help generalization.

We summarize the individual test-set predictions and their corresponding uncertainties in Figure~\ref{fig:result_condensed} (full version in Appendix, Figure~\ref{fig:results-full}).

\begin{figure}[h]
\centering
\vspace{-0.5em}
\includegraphics[width=1.0\linewidth]{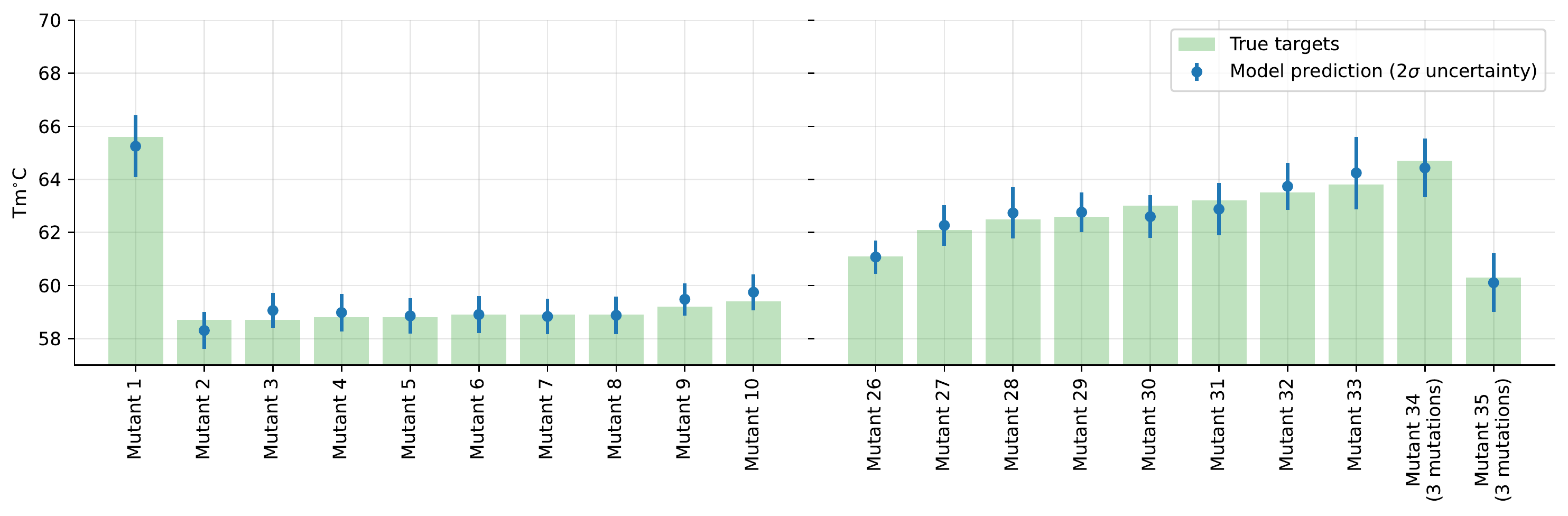}
\vspace{-2em}
\caption{A subset of melting temperature prediction and corresponding uncertainties on the test-set.}
\label{fig:result_condensed}
\end{figure}

Additionally, we train a model on a subset of the training-data comprising of only 80 samples all being single site mutations. While this restricted model does not perform particularly well in terms of our metric (resulting $\textrm{MAE} = 1.33$), it does great job in terms of uncertainties as shown in Figure~\ref{fig:result_1muts_condensed} (full version in Appendix, Figure~\ref{fig:results-1muts-full}). Note that this limited model is consistently underestimating the target, as it is unable to capture additive effects of mutations. 

\begin{figure}[h]
\centering
\vspace{-0.5em}
\includegraphics[width=1.0\linewidth]{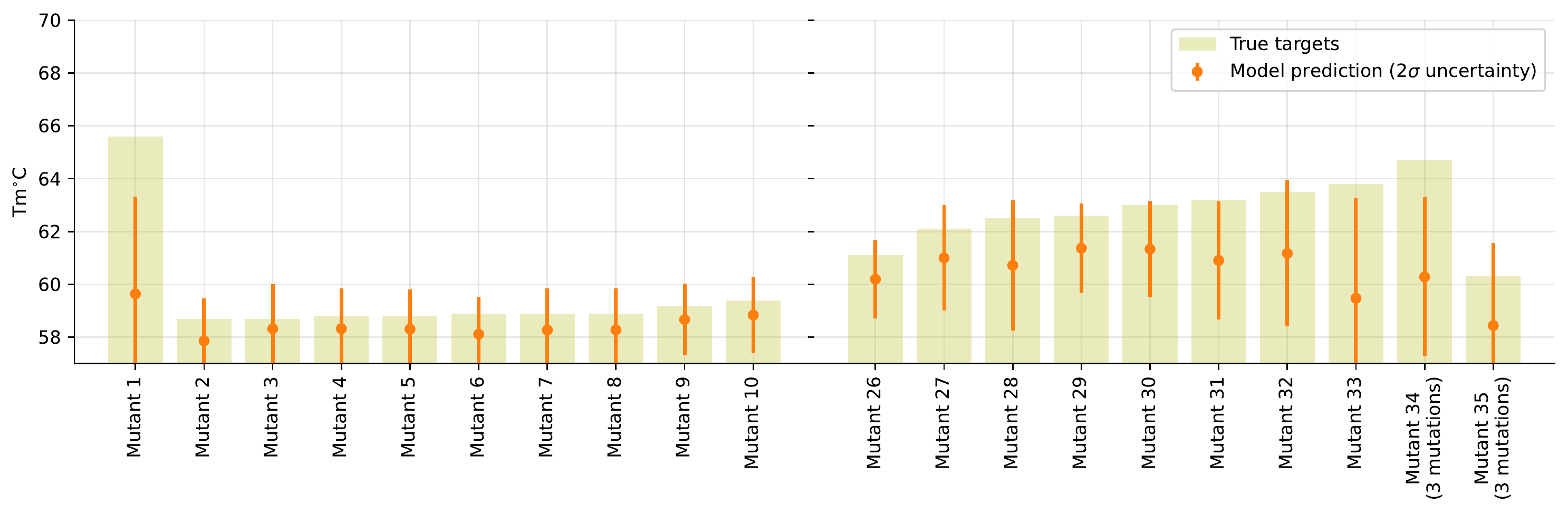}
\vspace{-2em}
\caption{A subset of melting temperature prediction and corresponding uncertainties on the test-set when model is trained only on the sigle-mutations $N{\textrm{train}}=80$}
\label{fig:result_1muts_condensed}
\end{figure}

%% file: tex/conclusions.tex
\section{Conclusions}
Utilization of machine learning driven techniques for navigating the drug design process towards optimized properties requires good compact representation of the molecule space of interest. We have demonstrated that sparsity promoting concentration of the larger pre-trained latent space provided by the protein language model, ESM-1b, leads to more robust estimate of a dedicated thermal stability optimization task for scFvs. We have proposed three variants of sparsity promoting effects through GP regression models integrating learned masks. In general all three models leads to improved predictive performance relative to a standard mean pooled feature representation. Even though our sparsity promoting models outperform our baseline model without mask on our final test data, our validation data indicates that the learned masks are sensitive to too aggressive sparsity when validation data is out-of-distribution of the training data. In fact, this makes sense as the sparse models exactly will seek to favor sparse representations given the training data at hand. Thus, if we seek to utilize the models solely for optimization in regions (residue positions) outside the support of previous seen data, care should be taken in utilizing the masks. From a Bayesian Optimization perspective, this type of evaluation would correspond to the explorative evaluation and thus the mean prediction evaluation is not suitable here. Instead of providing the mean prediction for evaluation we would seek opportunities to enrich the model support towards new regions e.g. through the upper-confidence-bound as acquisition function evaluation. Future work will examine the applicability sparsity promoting models in the context of steering explorative searches.

%% file: tex/appendix.tex
\appendix
\section{Appendix}

\begin{figure}[h]
\centering
\includegraphics[width=1.0\linewidth]{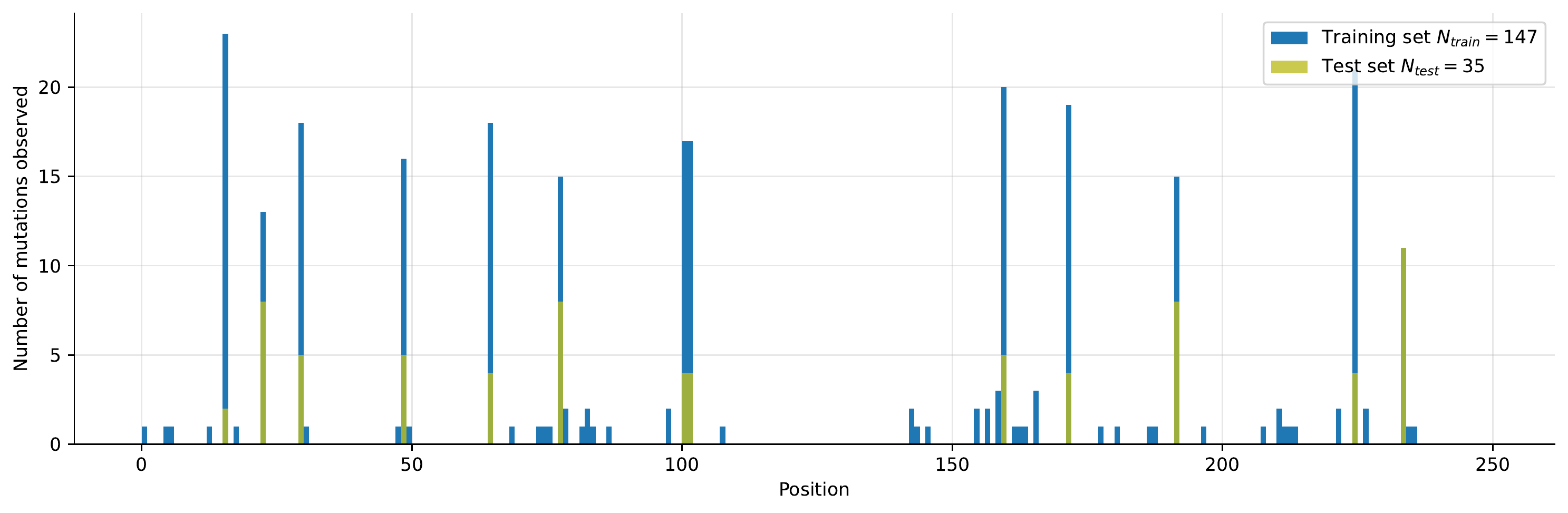}
\caption{Histogram of the mutations occurring at the respective positions in training and test set.}
\end{figure}

\begin{figure}[h]
\centering
\includegraphics[width=1.0\linewidth]{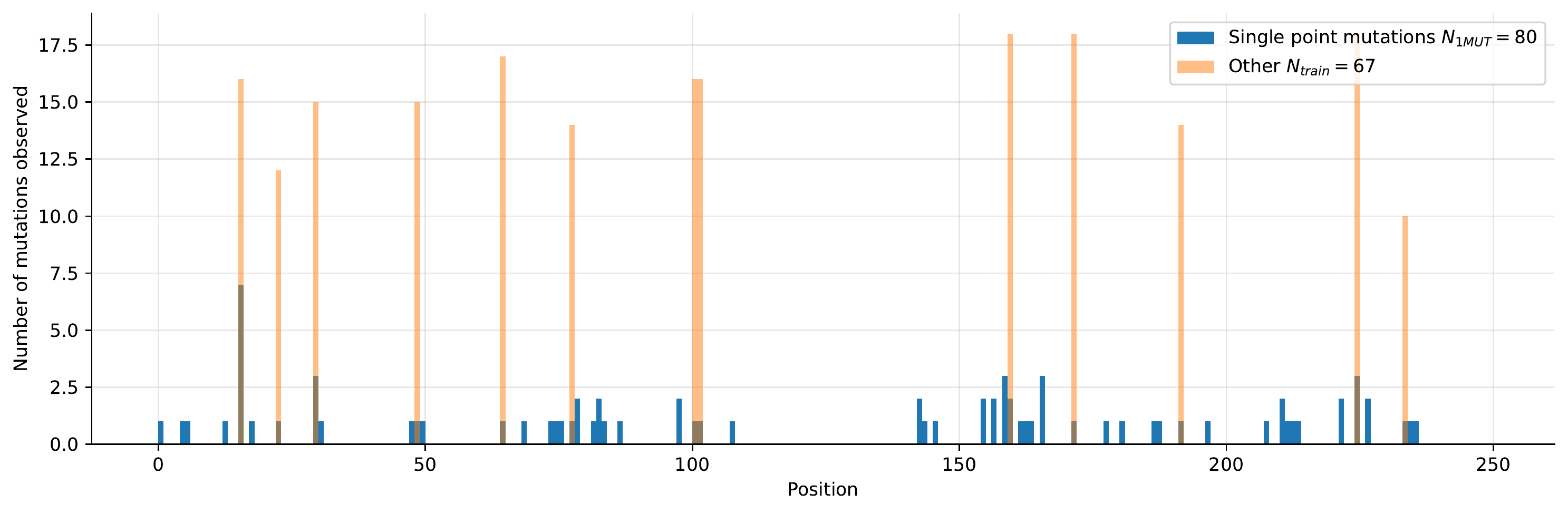}
\caption{Histogram of the single site mutations of the training set.}
\end{figure}

\begin{landscape}
\begin{figure}[!h]
\centering
\includegraphics[height=0.40\textheight]{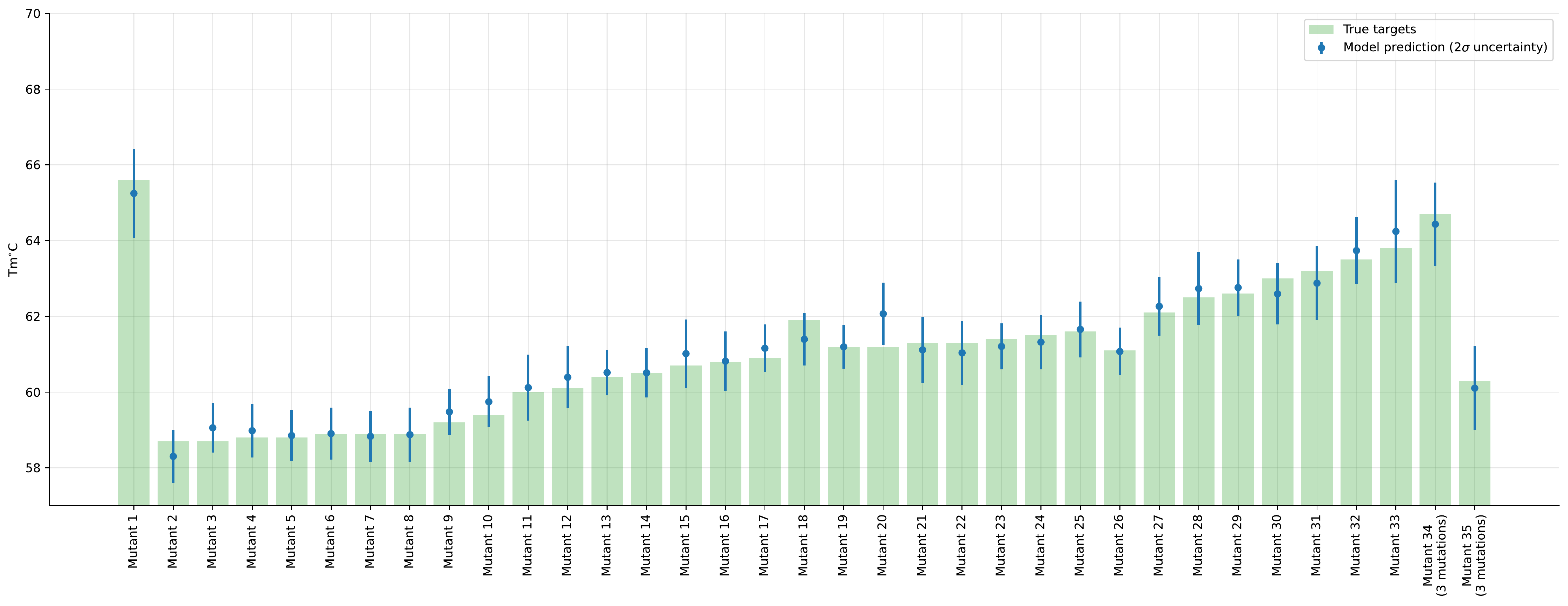}
\caption{Complete test-set melting temperature prediction and corresponding uncertainties.}
\label{fig:results-full}
\end{figure}

\begin{figure}[!h]
\centering
\includegraphics[height=0.40\textheight]{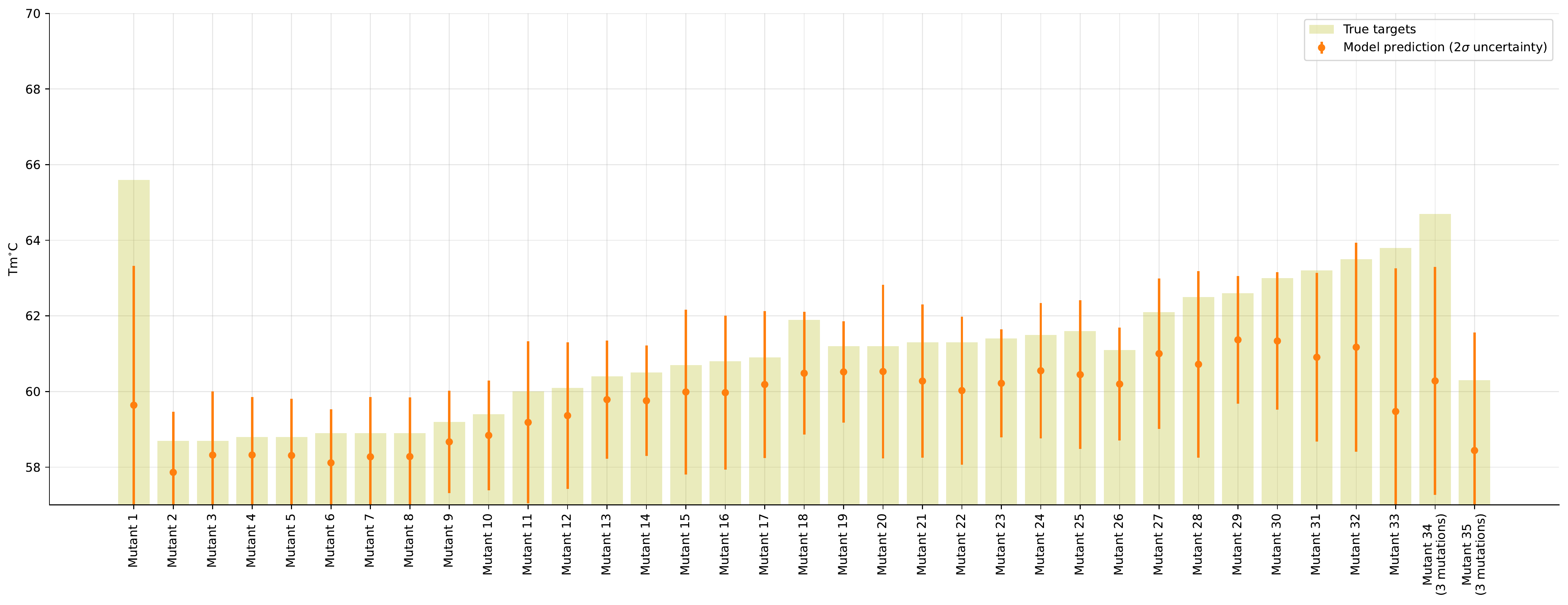}
\caption{Complete test-set melting temperature prediction and corresponding uncertainties.}
\label{fig:results-1muts-full}
\end{figure}
    
\end{landscape}